\documentclass{article}

\usepackage[preprint]{corl_2026} 

\usepackage{graphicx}
\usepackage{booktabs}
\usepackage{caption}
\usepackage{amsmath}
\usepackage{amssymb}
\usepackage{colortbl}
\usepackage{fontawesome5}

\definecolor{gaincolor}{rgb}{0.0,0.45,0.16}
\definecolor{dropcolor}{rgb}{0.72,0.0,0.0}
\definecolor{highlightcol}{rgb}{1.0,0.95,0.70}

\title{World Pilot: Steering Vision-Language-Action Models with World-Action Priors}

\newcommand{\equalcontrib}{\textsuperscript{*}}
\newcommand{\corrauth}{\textsuperscript{\faEnvelope}}

\author{
  \textbf{Zefu Lin}$^{1,2}$\equalcontrib\quad
  \textbf{Rongxu Cui}$^{3}$\equalcontrib\quad
  \textbf{Junjia Xu}$^{3}$\equalcontrib\quad
  \textbf{Xiaojuan Jin}$^{1}$ \quad
  \textbf{Wenling Li}$^{3}$ \\ 
  \textbf{Lue Fan}$^{1}$\corrauth\quad
  \textbf{Zhaoxiang Zhang}$^{1,2}$\corrauth
  \\
  $^1$ Institute of Automation, Chinese Academy of Sciences (CASIA)\\
  $^2$ Nanjing University \quad
  $^3$ Beihang University \\
  \texttt{\small \{linzefu2022, lue.fan\}@ia.ac.cn}
  \\
  \small \textsuperscript{*} Equal contribution. \quad
  \small \textsuperscript{\faEnvelope} Corresponding author.
  \\
  \\
  \small Project website: \href{https://world-pilot.github.io/}{\texttt{https://world-pilot.github.io/}}
}

\begin{document}
\maketitle

\vspace{-1.0em}
\begin{center}
\includegraphics[width=\textwidth]{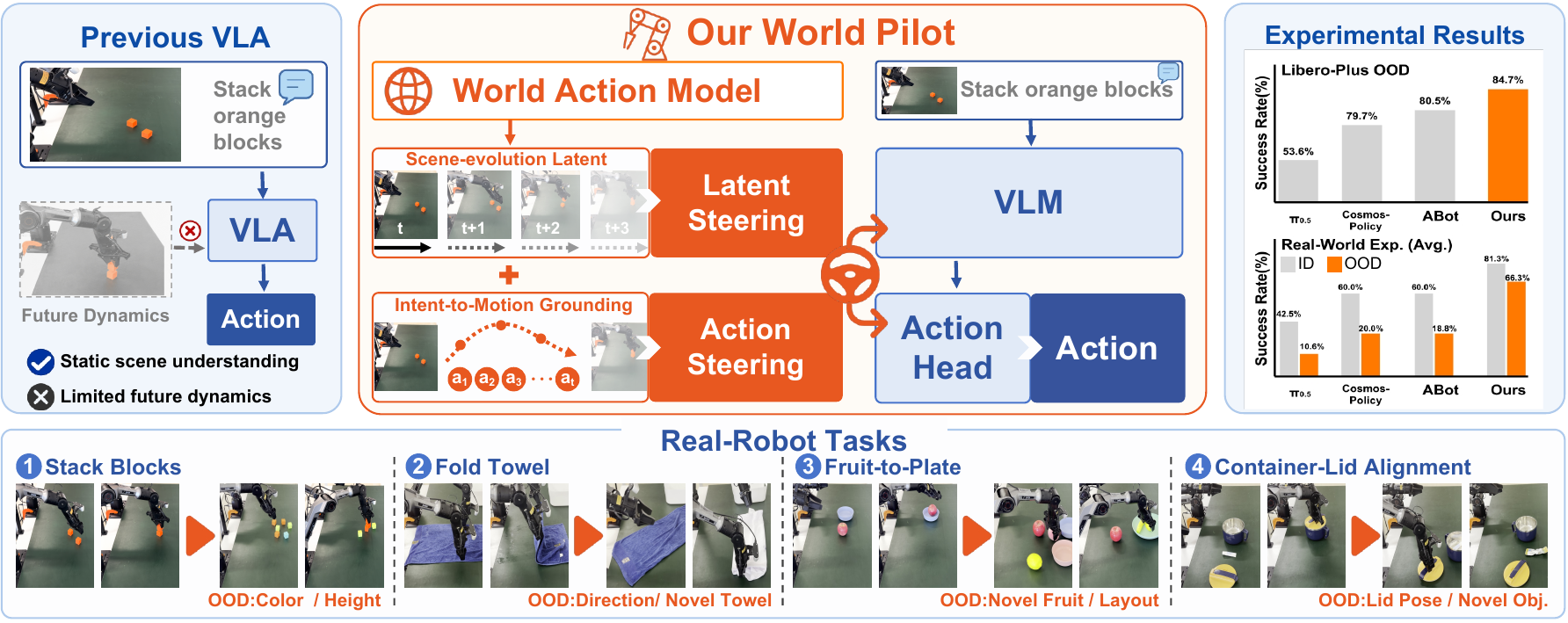}

\vspace{-0.5em}
\captionof{figure}{\textbf{World Pilot steers a VLA with priors from a World-Action Model.}
VLA methods generate actions from a VLM's encoding of the scene.
World Pilot adds two priors from a WAM into the decision chain, with \emph{Latent Steering} routing a scene-evolution latent into VLM hidden states and \emph{Action Steering} feeding a trajectory-level motion prior to the action generator.
This gives the VLA an anticipated view of the scene and a motion hint alongside its semantic conditioning.
World Pilot reaches state-of-the-art performance on LIBERO-Plus and real-robot tasks.}
\label{fig:teaser}
\end{center}


\begin{abstract}
Vision-Language-Action (VLA) models inherit semantic grounding from large-scale pretraining and perform competently across in-distribution manipulation tasks.
This grounding, however, is built on static image-text pairs, whereas manipulation is a continuous, contact-rich process whose dynamics this pretraining cannot capture.
We present World Pilot, a VLA framework that augments the policy with priors from a World-Action Model (WAM), routed into the decision chain through two complementary pathways. \emph{Latent Steering} conditions the perception layer on a scene-evolution latent, and \emph{Action Steering} supplies an anticipated trajectory as a motion prior to the action generator.
Together the two priors equip the VLA with an anticipated view of the scene and a trajectory-level motion hint alongside its semantic conditioning, and the scene-evolution prior remains effective even when supplied by a video-pretrained world model that has not been action-post-trained.
World Pilot attains a state-of-the-art Total success rate of 84.7\% on the LIBERO-Plus zero-shot OOD benchmark and the highest success rate on every real-robot setting across four manipulation tasks, with the largest margins under shifts in viewpoint, geometry, deformable state, and pose.

\end{abstract}



\section{Introduction}
\label{sec:introduction}
Vision-Language-Action (VLA) policies~\cite{v1,v2,v3,v4} inherit semantic grounding from the image-text pretraining of their VLM backbones and perform competently within the manipulation distribution on which they are fine-tuned~\cite{pi0.5,abot,li2024cogact}.
What such pretraining cannot supply is a model of how a scene evolves under action.
Image-text pairs are static~\cite{w19,w20,f10}, and the action generator downstream of the VLM consumes purely semantic hidden states with no internal account of the dynamics it must produce~\cite{Zhao-RSS-23,pi0,v14,v15,v16}.
Consistent with this gap, VLAs become fragile once viewpoint, geometry, or contact tolerance drifts away from the training distribution~\cite{v10,v11,v12,v13}.

Video pretraining is the natural complement.
Action-conditioned scene evolution is present in video by construction~\cite{w10,w11,w12}, and video-pretrained World-Action Models (WAMs) such as Cosmos Policy~\cite{cosmospolicy}, mimic-video~\cite{mimicvideo}, and DreamZero~\cite{dreamzero} acquire representations of scene dynamics that transfer broadly across embodiments and visual conditions~\cite{w1,w2,w3,w4,w5,w6,w7,w8,w9}.
Their outputs map onto exactly what VLAs lack: a scene-evolution latent describing how the visible state will change, and a coarse action-trajectory hypothesis sketching the actions whose effects the latent forecasts~\cite{f11,f12,f13}.
Because both predictions come from a shared encoder under joint training, they remain structurally aligned.
The two are therefore naturally complementary, with semantic grounding supplied by the VLA and scene dynamics supplied by the WAM.

Realizing this complementarity in practice, however, requires more than placing the two models side by side.
Whether the WAM's signals translate into a more capable policy depends on which signals to extract, in what form to carry them, and at which layers of the VLA to inject them, so that dynamics knowledge reaches the parts of the policy that need it without being diluted in transit.

We answer this question with \textbf{World Pilot}, a VLA framework that routes WAM outputs into the policy through two complementary pathways.
\emph{Latent Steering} injects the scene-evolution latent into VLM hidden states through a residual cross-attention update at the perception layer, supplying \emph{spatiotemporal dynamics anticipation}.
We route the latent rather than a decoded future image because pixel content carries action-irrelevant detail such as texture, lighting, background, and generation artifacts that dilute the dynamics structure the latent encodes directly~\cite{univla,stamo,lamlearn,wog}.
\emph{Action Steering} compresses the anticipated trajectory into a single prefix token at the flow-matching action generator, supplying \emph{intent-to-motion grounding} through a trajectory-level signal that biases generation toward the WAM's overall motion shape.
The single-token form leaves the generator free to commit to a specific continuous chunk informed by both the prior and the dynamics-enhanced hidden states.
The two priors enter at different layers because they carry different kinds of information, and both are additive.
Throughout fine-tuning the WAM is kept frozen, with gradient updates restricted to the VLA parameters and the lightweight fusion modules.
Both pathways therefore \emph{steer} the VLA with an existing world model rather than co-train a new one, and VLA fine-tuning never propagates back into the WAM to disturb its pretrained world prior.

The form and entry point of each prior are not freely interchangeable.
Several otherwise plausible alternatives, including a decoded future image in place of the latent, per-step trajectory tokens at the action generator, and flow-matching initialization from the WAM's trajectory, each tie the policy too tightly to a noisy intermediate output and forfeit part of the WAM's complementary dynamics signal.
Our ablations (Section~\ref{sec:ablations}) benchmark World Pilot against these alternatives under matched training conditions, and only World Pilot's specific configuration consistently converts the WAM's complementarity into measurable gain on the LIBERO-Plus OOD benchmark.

We evaluate World Pilot on LIBERO-Plus~\cite{libero-plus} and RoboCasa~\cite{robocasa}, and on four real-robot manipulation tasks.
World Pilot reaches a state-of-the-art Total success rate of 84.7\% on LIBERO-Plus and the highest success rate on every real-robot setting, while remaining competitive on RoboCasa.
Margins are largest under shifts in viewpoint, geometry, deformable state, and pose; ablations show that each pathway contributes independently and that the scene-evolution prior remains effective even when supplied by a video-pretrained world model that has not been action-post-trained.


\section{Related Work}
\label{sec:related_works}
\paragraph{Vision-Language-Action Models.}
Vision-Language-Action (VLA) policies attach an action generator to a Vision-Language Model (VLM) backbone~\cite{v17,v18,v19,v20}, producing continuous robot actions from visual observations and language instructions~\cite{v1,v2,v3,v4}.
Recent systems such as $\pi_{0.5}$~\cite{pi0.5}, ABot-M0~\cite{abot}, and CogACT~\cite{li2024cogact} achieve competent in-distribution performance on standard manipulation benchmarks.
Their conditioning, however, is built from image-text pretraining alone, with no representation of how the scene will evolve under actions, and they remain fragile under shifts in appearance, viewpoint, and physical interaction~\cite{v5,v6,v7,v8,v9}.

\paragraph{World-Action Models.}
World-Action Models (WAMs) such as Cosmos Policy~\cite{cosmospolicy}, mimic-video~\cite{mimicvideo}, and DreamZero~\cite{dreamzero} are pretrained on large-scale video sequences~\cite{w1,w2,w3,w4,w5}, learning the action-conditioned scene evolution and contact dynamics that image-text pretraining cannot capture.
Their video-pretrained representations transfer broadly across embodiments and visual conditions~\cite{w6,w7,w8,w9,w10,w11,w12}.
A natural design question is how to combine WAM-derived priors with a VLA's instruction-following pipeline~\cite{f1,f2,f3,f4,f5}, and prior work has explored several routes.
Motus~\cite{motus} and DreamVLA~\cite{dreamvla} jointly generate future images and actions in a unified framework, but the visual reconstruction loss pushes the action representation to absorb appearance details unrelated to control.
$\pi_{0.7}$~\cite{pi0.7} and VISTA~\cite{vista} use predicted future images or subgoal images to guide policy learning; pixel-space outputs encode appearance details such as texture, lighting, background, and generation artifacts that are largely irrelevant to action selection and dilute the control-relevant structure of the underlying world-model latent~\cite{f6,f7,f8,f9,univla,stamo,lamlearn}.
Being-H0.7~\cite{beingh0.7} and WoG~\cite{wog} pass world-model knowledge through latents or implicit features, reducing pixel-level information loss, but still rely on static future snapshots rather than continuous spatiotemporal evolution.
World Pilot instead routes two signals from a WAM into a VLA pipeline: a scene-evolution latent that conditions VLM hidden states through \emph{Latent Steering}, and an anticipated action trajectory that conditions the action generator through \emph{Action Steering}.


\section{Method}
\label{sec:method}
{\setlength{\textfloatsep}{8pt plus 2pt minus 2pt}%
\begin{figure}[t]
\centering
\includegraphics[width=\textwidth]{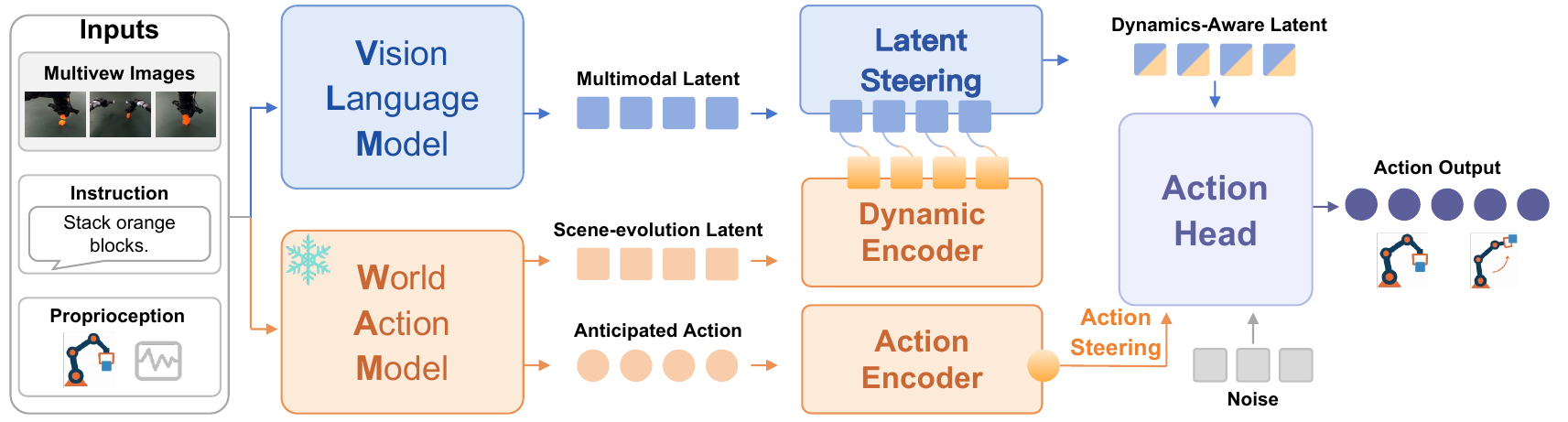}
\caption{\textbf{World Pilot architecture.}
A semantic pathway encodes images and language with a VLM into hidden states.
Two prior pathways from a World-Action Model enter the same decision chain, with \emph{Latent Steering} routing a scene-evolution latent into the VLM hidden states and \emph{Action Steering} compressing the anticipated trajectory into a prior token for the flow-matching action generator.}
\label{fig:framework}
\end{figure}}

\subsection{Problem Formulation}
\label{sec:problem_formulation}

We study robot manipulation policies conditioned on visual observations and natural-language instructions.
At each time step, the policy receives these inputs together with an optional proprioceptive state and predicts an action chunk $\mathbf{A}_t=(a_t,\ldots,a_{t+K-1})$ that controls the robot over a future horizon.
A standard Vision-Language-Action (VLA) policy encodes images and language with a Vision-Language Model (VLM) into multimodal hidden states, from which an action generator produces $\mathbf{A}_t$.
This pipeline inherits semantic grounding from image-text pretraining, but image-text pairs do not capture how a scene evolves under actions, and the action generator that follows operates on these purely semantic VLM hidden states.

World Pilot extends this pipeline with a video-pretrained World-Action Model (WAM) that, from the same inputs, jointly predicts a scene-evolution latent and a coarse action-trajectory hypothesis from a shared encoder, so the two outputs are structurally aligned, with the trajectory describing the actions whose effects the latent forecasts.
Let $\mathbf{O}_t$ denote the visual observation, $\ell$ the language instruction, and $\mathbf{q}_t$ the proprioceptive state when available.
The WAM branch returns $\mathbf{Z}^{w}_t$ and $\widetilde{\mathbf{A}}^{w}_t$, and World Pilot predicts the executable action chunk as
\begin{equation}
(\mathbf{Z}^{w}_t,\widetilde{\mathbf{A}}^{w}_t)=W_{\phi}(\mathbf{O}_t,\ell,\mathbf{q}_t), \qquad
\hat{\mathbf{A}}_{\theta,t}=\pi_{\theta}(\mathbf{O}_t,\ell,\mathbf{q}_t;\mathbf{Z}^{w}_t,\widetilde{\mathbf{A}}^{w}_t),
\end{equation}
where $\mathbf{Z}^{w}_t$ is the scene-evolution latent, $\widetilde{\mathbf{A}}^{w}_t$ is the anticipated action trajectory used as a motion prior, and $\hat{\mathbf{A}}_{\theta,t}$ is the action chunk produced by World Pilot.

Two pathways route the WAM outputs into the policy (Fig.~\ref{fig:framework}).
A semantic backbone first encodes the current images and instruction into VLM hidden states $\mathbf{H}_t$.
\emph{Latent Steering} (Section~\ref{sec:latent_steering}) conditions $\mathbf{H}_t$ on $\mathbf{Z}^{w}_t$ to produce dynamics-enhanced hidden states $\bar{\mathbf{H}}_t$, supplying spatiotemporal dynamics anticipation at the perception layer.
\emph{Action Steering} (Section~\ref{sec:action_steering}) encodes $\widetilde{\mathbf{A}}^{w}_t$ into a single trajectory-level prior token $\mathbf{s}^{w}_t$, supplying intent-to-motion grounding at the action-generation layer.
Both pathways are additive: Latent Steering adds a residual to $\mathbf{H}_t$ that preserves the token sequence, and Action Steering inserts a single prefix token into the generator without altering its denoising recurrence, so each pathway can be ablated independently as Section~\ref{sec:result} confirms.

\subsection{Latent Steering}
\label{sec:latent_steering}

Latent Steering enriches VLM hidden states with cues about anticipated scene evolution.
The scene-evolution latent $\mathbf{Z}^{w}_t$ carries compact information about predicted object motion, contact outcomes, and local state changes.
We use this latent rather than a decoded future image because pixel content carries action-irrelevant detail such as texture, lighting, background, and generation artifacts that dilute the dynamics structure the latent encodes directly~\cite{univla,stamo,wog}.

Given the current observation and instruction, the WAM predicts $\mathbf{Z}^{w}_t$ as a per-view latent representation of the future visual state.
Concretely, the WAM encodes $\mathbf{O}_t$ with a VAE and denoises it via a Diffusion Transformer (DiT), yielding $\mathbf{Z}^{w}_t$.
World Pilot projects this latent through a dynamics encoder $f_{\mathrm{dyn}}$ and adds a temporal embedding $\boldsymbol{\rho}_{\mathrm{fut}}$ that marks the tokens as future-scene tokens, giving $\mathbf{D}^{w}_t=f_{\mathrm{dyn}}(\mathbf{Z}^{w}_t)+\boldsymbol{\rho}_{\mathrm{fut}}$; without this tag, the prior's contribution diminishes empirically.
Let $\mathbf{H}_t\in\mathbb{R}^{L\times d}$ denote the VLM hidden states.
The Latent Steering block applies cross-attention from $\mathbf{H}_t$ to $\mathbf{D}^{w}_t$ and adds the result back as a residual,
\begin{equation}
\bar{\mathbf{H}}_t = \mathbf{H}_t + \operatorname{CrossAttn}(\mathbf{H}_t,\mathbf{D}^{w}_t).
\end{equation}
Cross-attention lets each VLM token attend selectively to the parts of $\mathbf{D}^{w}_t$ most relevant to its spatial region, rather than receiving a single global modulation.
The residual form preserves the original VLM token order and hidden-state structure, so $\bar{\mathbf{H}}_t$ feeds directly into the standard VLA action-generation path with no further adaptation or downstream interface change.

\subsection{Action Steering}
\label{sec:action_steering}

Action Steering supplies the action generator with a soft trajectory-level context derived from $\widetilde{\mathbf{A}}^{w}_t$.
This context guides generation rather than replacing it, and the executed trajectory remains the output of the VLA action generator under standard action supervision.

The WAM produces $\widetilde{\mathbf{A}}^{w}_t$ with a horizon and action dimension that depend on the task.
World Pilot aligns this trajectory to the VLA action horizon $K$ by resampling and encodes the result with an action encoder $f_{\mathrm{act}}$ into a single prior token $\mathbf{s}^{w}_t=f_{\mathrm{act}}(\mathrm{Align}_{K}(\widetilde{\mathbf{A}}^{w}_t))$.
A single token summarizes the trajectory's overall shape rather than pinning generation to per-step targets, leaving the generator free to commit to a specific continuous chunk that reflects both the prior and the dynamics-enhanced hidden states.
Per-step conditioning, in contrast, ties each output step to the corresponding WAM step, which we find empirically to be less robust when the WAM trajectory is approximate.

The flow-matching action generator denoises a noisy trajectory $\mathbf{X}_{\tau,t}$ at flow time $\tau$ toward the clean action chunk.
World Pilot extends its input to $[\mathbf{u}_t;\mathbf{s}^{w}_t;\mathbf{Q}_t;\mathbf{X}_{\tau,t}]$, where $\mathbf{u}_t$ is the optional state token and $\mathbf{Q}_t$ are learned future-query tokens.
The dynamics-enhanced VLM hidden states $\bar{\mathbf{H}}_t$ provide the cross-attention condition.
$\mathbf{s}^{w}_t$ enters as a prefix rather than as part of the noisy trajectory, so it conditions the denoising recurrence through self-attention without itself being denoised.
Table~\ref{tab:ablation_action_form} compares this single-token form against three alternative ways of feeding $\widetilde{\mathbf{A}}^{w}_t$ to the generator and shows that the encoded single token attains the highest success rate.

\subsection{Policy Training}
\label{sec:policy_training}

Each training sample provides the observation, language instruction, optional proprioceptive state, and expert action chunk $\mathbf{A}^{\star}_t$.
Throughout fine-tuning the WAM $W_{\phi}$ is kept frozen, with gradient updates restricted to the VLA-side parameters $\theta$ (the VLM backbone, the dynamics encoder $f_{\mathrm{dyn}}$ and Latent Steering cross-attention, the action encoder $f_{\mathrm{act}}$, and the flow-matching action generator).
We therefore treat the WAM as an external prior model rather than a component to be co-trained, which is the sense in which World Pilot \emph{steers} the VLA with an existing world model: VLA fine-tuning does not propagate gradients back into the WAM, and its forward pass can be precomputed and cached so that it is excluded from the inner training loop.
At inference, both the VLA and the WAM run online and produce the priors from the live observation at every decision step, and the fusion paths in Sections~\ref{sec:latent_steering}--\ref{sec:action_steering} consume identically shaped priors at training and inference, so the learned fusion behavior transfers directly to online execution.

Following ABot-M0~\cite{abot}, we adopt the clean-action parameterization of the flow-matching action generator, which is equivalent to a reweighted velocity-space objective induced by the action-to-velocity transformation.
The parameterization keeps the supervision target equal to the expert chunk $\mathbf{A}^{\star}_t$, so the WAM priors enter only through the conditioning path and need no separate prior loss.
Given Gaussian noise $\boldsymbol{\epsilon}$ and a sampled flow time $\tau$, the noisy trajectory is $\mathbf{X}_{\tau,t}=\tau\mathbf{A}^{\star}_t+(1-\tau)\boldsymbol{\epsilon}$, and the action generator predicts a clean action chunk
\begin{equation}
\hat{\mathbf{A}}_{\theta,t}
=
g_{\theta}\!\left(\mathbf{X}_{\tau,t},\tau,\mathbf{u}_t,\mathbf{s}^{w}_t,\mathbf{Q}_t \mid \bar{\mathbf{H}}_t\right).
\end{equation}
The training objective is
\begin{equation}
\mathcal{L}_{\text{World Pilot}}=
\mathbb{E}_{\tau,\boldsymbol{\epsilon}}\!\left[
w(\tau)\left\|\hat{\mathbf{A}}_{\theta,t}-\mathbf{A}^{\star}_t\right\|_2^2
\right], \qquad
w(\tau)=\frac{1}{(1-\tau)^2},
\end{equation}
where $w(\tau)$ implements the equivalent velocity-space loss under this parameterization.
Optimizing this objective end-to-end teaches World Pilot how to use the world priors provided by Latent Steering and Action Steering to guide the action generator toward lower-error actions.


\section{Experimental Results}
\label{sec:result}
\subsection{Main Experiments}
\label{sec:setup_and_main}

We build World Pilot on the ABot-M0~\cite{abot}, with Qwen3-VL~\cite{qwen3} as the VLM backbone and a DiT-based flow-matching action head, and use Cosmos Policy~\cite{cosmospolicy} as the WAM with 5-step denoising.
WAM outputs are precomputed during training and run online at evaluation.
We apply dropout with rate $0.3$ to the WAM conditions $\mathbf{D}^{w}_t$ and $\mathbf{s}^{w}_t$ to prevent the policy from over-relying on the priors.
We fine-tune World Pilot on 8 RTX PRO 6000 GPUs and report success rate.

We evaluate on two simulation benchmarks.
LIBERO-Plus~\cite{libero-plus} is an OOD suite of 10{,}030 perturbed tasks built on LIBERO~\cite{libero} that covers seven axes of perturbation (background, camera, language, light, layout, robot, noise).
Models are trained only on LIBERO and evaluated zero-shot on the perturbations, with Total reporting the success rate averaged over all perturbed tasks.
RoboCasa~\cite{robocasa} emphasizes long-horizon manipulation in everyday kitchen scenes.

{\setlength{\textfloatsep}{8pt plus 2pt minus 2pt}%
\begin{table}[t]
\caption{\textbf{Simulation results on LIBERO, LIBERO-Plus, and RoboCasa.}
All LIBERO-Plus numbers come from training on LIBERO only and evaluating zero-shot on its OOD perturbations.
The LIBERO-Plus numbers for Cosmos Policy~\cite{cosmospolicy} and DreamVLA~\cite{dreamvla} are our own runs, as is the RoboCasa number for ABot-M0~\cite{abot}, and the remaining LIBERO-Plus baselines are taken from ABot-M0 and Being-H0.7~\cite{beingh0.7}.
We rerun ABot-M0 on RoboCasa because the ABot-M0 paper reports RoboCasa on the GR1 split rather than the original benchmark used here.}
\label{tab:main_results}
\centering
\small
\setlength{\tabcolsep}{3.2pt}
\renewcommand{\arraystretch}{1.12}
\resizebox{\textwidth}{!}{%
\begin{tabular}{lcccccccc>{\columncolor{highlightcol}}cc}
\toprule
Method & LIBERO & \multicolumn{8}{c}{LIBERO-Plus} & RoboCasa \\
\cmidrule(lr){3-10}
&  & Camera & Robot & Language & Light & Background & Noise & Layout & Total &  \\
\midrule
OpenVLA~\cite{openvla} & 84.7 & 0.8 & 3.5 & 23.0 & 8.1 & 34.8 & 15.2 & 28.5 & 15.6 & -- \\
WorldVLA~\cite{worldVLA} & 81.8 & 0.1 & 27.9 & 41.6 & 43.7 & 17.1 & 10.9 & 38.0 & 25.0 & -- \\
UniVLA~\cite{univla} & 95.2 & 1.8 & 46.2 & 69.6 & 69.0 & 81.0 & 21.2 & 31.9 & 42.9 & -- \\
$\pi_0$~\cite{pi0} & 94.4 & 13.8 & 6.0 & 58.8 & 85.0 & 81.4 & 79.0 & 68.9 & 53.6 & 42.4 \\
$\pi_{0.5}$~\cite{pi0.5} & 96.9 & -- & -- & -- & -- & -- & -- & -- & 77.4 & 41.4 \\
RIPT-VLA & 93.6 & 55.2 & 31.2 & 77.6 & 88.4 & \underline{91.6} & 73.5 & 74.2 & 68.4 & -- \\
DreamVLA~\cite{dreamvla} & 97.5 & 26.2 & 17.6 & 67.0 & 77.5 & 71.5 & 53.6 & 43.5 & 48.9 & -- \\
Being-H0.7~\cite{beingh0.7} & \textbf{99.2} & -- & -- & -- & -- & -- & -- & -- & \underline{82.1} & 62.1 \\
Cosmos Policy~\cite{cosmospolicy} & 98.5 & \underline{69.6} & 51.0 & \textbf{89.6} & \underline{97.7} & 85.7 & \underline{87.3} & \textbf{83.7} & 79.7 & \textbf{67.1} \\
ABot-M0~\cite{abot} & \underline{98.6} & 60.4 & \textbf{67.9} & 86.4 & 96.2 & \underline{91.6} & 86.4 & \underline{82.6} & 80.5 & 54.0 \\
\midrule
\textbf{World Pilot (Ours)} & 98.5 & \textbf{82.8} & \underline{60.6} & \underline{87.2} & \textbf{98.6} & \textbf{96.4} & \textbf{93.6} & 80.5 & \textbf{84.7} & \underline{65.5} \\
\bottomrule
\end{tabular}
}
\end{table}}

World Pilot reaches the highest Total success rate on LIBERO-Plus~\cite{libero-plus} at 84.7\% averaged over three random seeds, a 2.6-point margin over the strongest reported baseline (Table~\ref{tab:main_results}), and leads on Camera, Light, Background, and Noise while placing close behind the strongest baselines on Language, Robot, and Layout.
On the appearance axes (Light, Background, Noise), World Pilot leads on all three, consistent with image-text pretraining at the VLM and video pretraining at the WAM both contributing appearance robustness.
On Camera, World Pilot reaches 82.8 (+13.2 over the next baseline), the largest per-axis gain in the table; the WAM's video pretraining covers diverse camera poses, and the scene-evolution latent carries this coverage into the policy, narrowing the gap that pretraining leaves open.
LIBERO-Plus reports that a high Language score reflects insensitivity to instruction perturbations rather than robustness~\cite{libero-plus}, so we read this column as a sanity check.
We treat Total as the primary indicator of broad OOD robustness, since the perturbation a deployed scene presents is unknown and Total aggregates over all 10{,}030 perturbed tasks.
On LIBERO~\cite{libero} itself, recent strong baselines already sit above 98\% with little headroom, so World Pilot's gains concentrate on the OOD axes.
On RoboCasa, World Pilot is competitive with the strongest reported baseline, so the same conditioning design carries over to long-horizon kitchen tasks.

\subsection{Real-World Experiments}
\label{sec:real_world}

{\setlength{\textfloatsep}{8pt plus 2pt minus 2pt}%
\begin{figure}[t]
\centering
\includegraphics[width=\textwidth]{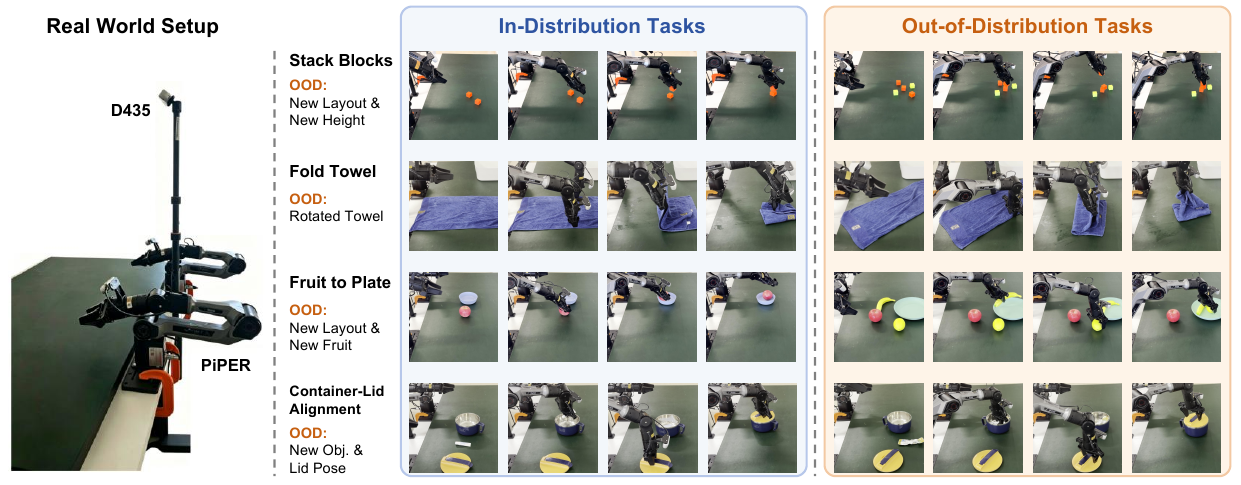}
\caption{\textbf{Real-robot evaluation setup and task scenes.}
The robot platform (left), in-distribution scenes matching the training conditions (middle), and out-of-distribution scenes (right) under changes in appearance, geometry, deformable state, or pose.}
\label{fig:real_robot}
\end{figure}}

The platform and the ID/OOD scenes are illustrated in Fig.~\ref{fig:real_robot}.
We compare World Pilot with ABot-M0~\cite{abot}, Cosmos Policy~\cite{cosmospolicy}, and $\pi_{0.5}$~\cite{pi0.5} on the same robot-arm platform and RGB inputs, across four manipulation tasks (stacking blocks, folding towels, placing fruit on a plate, and container-lid alignment), each with one ID and two OOD variants that change geometry, deformable state, or pose.
For each task we collect 100 ID teleoperated demonstrations, fine-tune all methods for 10{,}000 steps under a matched optimizer, batch size, and learning-rate schedule, and run 20 trials per task setting per method, scoring a trial as successful if the robot reaches the final state specified by the language instruction within the allowed time.
We read individual cell differences below 10 percentage points as within trial-level variance and rely on consistent direction across the 12 task-setting cells.

{\setlength{\textfloatsep}{8pt plus 2pt minus 2pt}%
\begin{table}[t]
\centering
\caption{
\textbf{Real-robot success rates on four manipulation tasks.}
Each task has one in-distribution (ID) setting that matches training and two out-of-distribution (OOD) variants that perturb appearance, geometry, deformable state, or pose, and success is measured over 20 trials per setting.
Parenthesized red values give the absolute drop from the corresponding ID setting.
}
\label{tab:real_robot_static_physical_transition}
\small
\setlength{\tabcolsep}{2.4pt}
\renewcommand{\arraystretch}{1.14}
\resizebox{\linewidth}{!}{%
\begin{tabular}{@{}lcccccccccccc@{}}
\toprule
Method
& \multicolumn{3}{c}{Stack Blocks}
& \multicolumn{3}{c}{Fold Towel}
& \multicolumn{3}{c}{Fruit-to-Plate}
& \multicolumn{3}{c}{Container-Lid Alignment} \\
\cmidrule(lr){2-4}
\cmidrule(lr){5-7}
\cmidrule(lr){8-10}
\cmidrule(lr){11-13}
& ID & Color & Height
& ID & Direction & Novel Towel
& ID & Novel Fruit & Layout
& ID & Novel Object & Lid Pose \\
\midrule
$\pi_{0.5}$~\cite{pi0.5}
& 40 & 15 {\color{dropcolor}{(-25)}} & 0 {\color{dropcolor}{(-40)}}
& 55 & 25 {\color{dropcolor}{(-30)}} & 10 {\color{dropcolor}{(-45)}}
& 35 & 10 {\color{dropcolor}{(-25)}} & 5 {\color{dropcolor}{(-30)}}
& 40 & 15 {\color{dropcolor}{(-25)}} & 5 {\color{dropcolor}{(-35)}} \\

ABot-M0~\cite{abot}
& 60 & 25 {\color{dropcolor}{(-35)}} & 10 {\color{dropcolor}{(-50)}}
& 50 & 20 {\color{dropcolor}{(-30)}} & 5 {\color{dropcolor}{(-45)}}
& 65 & 30 {\color{dropcolor}{(-35)}} & 15 {\color{dropcolor}{(-50)}}
& 65 & 30 {\color{dropcolor}{(-35)}} & 15 {\color{dropcolor}{(-50)}} \\

Cosmos Policy~\cite{cosmospolicy}
& 65 & 30 {\color{dropcolor}{(-35)}} & 15 {\color{dropcolor}{(-50)}}
& 45 & 15 {\color{dropcolor}{(-30)}} & 10 {\color{dropcolor}{(-35)}}
& 70 & 35 {\color{dropcolor}{(-35)}} & 20 {\color{dropcolor}{(-50)}}
& 60 & 25 {\color{dropcolor}{(-35)}} & 10 {\color{dropcolor}{(-50)}} \\

\textbf{World Pilot (Ours)}
& \textbf{70} & \textbf{55} {\color{dropcolor}{\textbf{(-15)}}} & \textbf{50} {\color{dropcolor}{\textbf{(-20)}}}
& \textbf{85} & \textbf{75} {\color{dropcolor}{\textbf{(-10)}}} & \textbf{70} {\color{dropcolor}{\textbf{(-15)}}}
& \textbf{90} & \textbf{75} {\color{dropcolor}{\textbf{(-15)}}} & \textbf{70} {\color{dropcolor}{\textbf{(-20)}}}
& \textbf{80} & \textbf{70} {\color{dropcolor}{\textbf{(-10)}}} & \textbf{65} {\color{dropcolor}{\textbf{(-15)}}} \\
\bottomrule
\end{tabular}
}
\vspace{0.4em}
\begin{minipage}{0.98\linewidth}
\footnotesize
\textbf{OOD settings.}
\textit{Stack Blocks}, block color and stacking height.
\textit{Fold Towel}, towel direction and towel instance.
\textit{Fruit-to-Plate}, fruit category and fruit/plate layout.
\textit{Container-Lid Alignment}, object category and lid pose, where success requires the lid to be aligned with the container rim and fully closed.
\end{minipage}
\end{table}}

World Pilot attains the highest success rate on every setting in Table~\ref{tab:real_robot_static_physical_transition}, with the largest margins under OOD perturbations.
World Pilot's ID-to-OOD drop stays within 20 absolute points, while other baselines drop by 25 to 50.
Container-lid alignment is the most stringent setting, requiring tight geometric tolerance for closure; under OOD pose and object changes World Pilot succeeds in 13 to 14 of 20 trials, while no baseline exceeds 6.
Both priors thus remain effective when the object's geometry, pose, or appearance moves outside the training distribution, providing trajectory-level conditioning and anticipated scene-state cues that VLM hidden states do not carry.

\subsection{Ablations}
\label{sec:ablations}

We organize ablations by pathway, with \emph{Latent Steering} on the perception side and \emph{Action Steering} on the action-generator side.
We first show that each pathway contributes, then probe the source and form of Latent Steering's prior, and finally vary how the action-trajectory prior enters the generator.

\textbf{Each pathway contributes.}
We first isolate the two pathways on LIBERO-Plus~\cite{libero-plus} (Table~\ref{tab:ablation_1}).
Latent Steering alone reaches 83.7\% (+3.2 over the 80.5\% ABot-M0~\cite{abot} baseline), Action Steering alone reaches 83.1\% (+2.6), and combining the two pathways gives the strongest result at 84.7\%, indicating that anticipated scene dynamics and trajectory-level priors contribute complementary signals beyond the VLM's semantic representation.

\textbf{Latent Steering: the world prior is already present before action fine-tuning.}
We test whether the world prior Latent Steering consumes is already supplied by a world model that produces only future-scene predictions, namely Cosmos-Predict~\cite{cosmos-predict}, or whether it requires the further action post-training that adapts Cosmos-Predict into Cosmos Policy~\cite{cosmospolicy}.
Cosmos-Predict is pretrained on large-scale, filtered, VLM-captioned video and image data covering Physical AI scenes such as driving, robot manipulation, human activity, navigation, natural physical dynamics, first-person views, and synthetic rendering, so its scene-evolution latent already encodes broadly transferable dynamics structure before any action-side adaptation.
We take Cosmos-Predict in scene-prediction-only mode, route its scene-evolution latent through Latent Steering with Action Steering disabled, and additionally evaluate on RoboTwin2.0 (clean)~\cite{robotwin} to test how broadly the world prior transfers.
This world-model-only signal still improves over ABot-M0 on every benchmark (Table~\ref{tab:ablation_frozen_wam}), reaching 82.6 (+2.1) on LIBERO-Plus~\cite{libero-plus}, 62.7 (+8.7) on RoboCasa, and 85.3 (+4.1) on RoboTwin2.0 (clean).
The action post-training that adapts Cosmos-Predict into Cosmos Policy~\cite{cosmospolicy} further sharpens the signal (the Cosmos-Policy-based counterpart in Table~\ref{tab:ablation_1} reaches 83.7\% on LIBERO-Plus with Latent Steering only, +1.1 over the Cosmos-Predict-only setting under matched projection head, dropout, and training schedule), but the prior takes effect even without it.

\textbf{Latent Steering: latent injection over decoded future images.}
Given that the latent prior is already useful from world-model pretraining alone, we ask in what form it should enter Latent Steering (Table~\ref{tab:ablation_2}), evaluating 1-step, 3-step, and 5-step latents (taken from intermediate Cosmos denoising states) together with a fully decoded future image passed through the VLA's image encoder.
Latent injection is stable across denoising depths, with the three latent variants falling within 0.2 points of each other (84.5 to 84.7\%), since World Pilot relies on state-transition cues and local dynamics structure encoded in the latent rather than on pixel-level realism.
Replacing the latent with a fully decoded future image instead lowers Total to 83.5\% (a 1.2-point drop), as pixel-level decoding adds visual artifacts and dilutes the dynamics structure.

\textbf{Action Steering: how the trajectory prior conditions the generator.}
We vary how the trajectory $\widetilde{\mathbf{A}}^{w}_t$ enters the action generator (Table~\ref{tab:ablation_action_form}).
We compare the single trajectory-level prior token against three alternatives: per-step encoded tokens, flow-matching initialization from $\widetilde{\mathbf{A}}^{w}_t$, and use of the raw $\widetilde{\mathbf{A}}^{w}_t$ as the action prior.
The single-token form gives the strongest result at 84.7\%.
Per-step tokens (83.6\%) and the raw trajectory (83.0\%) pin generation to noisy step-level signals, propagating WAM trajectory noise and compounding errors across the chunk.
Flow-matching initialization recovers part of this gap (84.1\%) but ties the final output to the WAM's action quality, leaving the generator less room to correct the prior with VLA-side cues.
Compressing the trajectory into a single conditioning token keeps the prior as guidance while the generator commits to a chunk that reflects both the prior and the dynamics-enhanced hidden states.

\setlength{\floatsep}{4pt plus 1pt minus 1pt}
{\setlength{\textfloatsep}{8pt plus 2pt minus 2pt}%
\begin{table}[t]
\begin{minipage}[t]{0.48\textwidth}
\centering
\caption{\textbf{Contribution of each prior pathway on LIBERO-Plus.} Each pathway is enabled individually in isolation and then evaluated in combination, and green values mark absolute gains over the ABot-M0 baseline.}
\label{tab:ablation_1}
\small
\setlength{\tabcolsep}{8pt}
\renewcommand{\arraystretch}{1.18}
\begin{tabular}{lc}
\toprule
\textbf{Variant} & \textbf{Success (\%)} \\
\midrule
ABot-M0~\cite{abot} (baseline) & 80.5 \\
Latent Steering only & 83.7 {\color{gaincolor}{(+3.2)}} \\
Action Steering only & 83.1 {\color{gaincolor}{(+2.6)}} \\
Full World Pilot & \textbf{84.7} {\color{gaincolor}{(+4.2)}} \\
\bottomrule
\end{tabular}
\end{minipage}
\hfill
\begin{minipage}[t]{0.48\textwidth}
\centering
\caption{\textbf{World-model-only prior transfer.} The WAM is replaced by Cosmos-Predict~\cite{cosmos-predict}, which has not been action-post-trained and produces only future latents, with only Latent Steering (LS) active. RoboTwin2.0 results are reported on the \emph{clean} split.}
\label{tab:ablation_frozen_wam}
\small
\setlength{\tabcolsep}{4pt}
\renewcommand{\arraystretch}{1.18}
\resizebox{\linewidth}{!}{%
\begin{tabular}{lcc}
\toprule
\textbf{Benchmark} & \textbf{ABot-M0} & \textbf{+ LS (world model)} \\
\midrule
LIBERO-Plus           & 80.5 & \textbf{82.6}\,{\color{gaincolor}{\scriptsize(+2.1)}} \\
RoboCasa              & 54.0 & \textbf{62.7}\,{\color{gaincolor}{\scriptsize(+8.7)}} \\
RoboTwin2.0      & 81.2 & \textbf{85.3}\,{\color{gaincolor}{\scriptsize(+4.1)}} \\
\bottomrule
\end{tabular}}
\end{minipage}

\par\vspace{2pt}
\begin{minipage}[t]{0.48\textwidth}
\centering
\caption{\textbf{Future-scene representation on LIBERO-Plus.} Latent rows take the WAM cue at a Cosmos denoising step, and the decoded variant replaces the latent with a decoded future image to test whether pixel-space realism helps.}
\label{tab:ablation_2}
\small
\setlength{\tabcolsep}{8pt}
\renewcommand{\arraystretch}{1.18}
\begin{tabular}{lc}
\toprule
\textbf{Future Information} & \textbf{Success (\%)} \\
\midrule
Future latent (1 step)  & 84.6 \\
Future latent (3 steps) & 84.5 \\
Future latent (5 steps) & \textbf{84.7} \\
Decoded future image    & 83.5 \\
\bottomrule
\end{tabular}
\end{minipage}
\hfill
\begin{minipage}[t]{0.48\textwidth}
\centering
\caption{\textbf{Action-prior form on LIBERO-Plus.} Four ways of feeding the WAM trajectory $\widetilde{\mathbf{A}}^{w}_t$ to the flow-matching generator are compared, varying granularity and entry point, where \emph{Ours} marks World Pilot's default setting.}
\label{tab:ablation_action_form}
\small
\setlength{\tabcolsep}{8pt}
\renewcommand{\arraystretch}{1.18}
\begin{tabular}{lc}
\toprule
\textbf{Action Prior Form} & \textbf{Success (\%)} \\
\midrule
Single encoded token (Ours)             & \textbf{84.7} \\
Per-step encoded tokens                 & 83.6 \\
Flow init.\ from $\widetilde{\mathbf{A}}^{w}_t$ & 84.1 \\
Raw $\widetilde{\mathbf{A}}^{w}_t$              & 83.0 \\
\bottomrule
\end{tabular}
\end{minipage}
\end{table}}


\section{Conclusion and Limitations}
\label{sec:conclusion}
We propose a training recipe that augments VLA policy learning with priors from a World-Action Model (WAM), routed through \emph{Latent Steering} on the perception side and \emph{Action Steering} on the action-generator side.
We instantiate this recipe as World Pilot, which attains state-of-the-art performance on LIBERO-Plus~\cite{libero-plus} and the highest success rate on every real-robot setting.

\textbf{Limitations.}
World Pilot inherits its WAM's coverage, so when test scenes fall outside the WAM's video pretraining distribution, both priors degrade and the gains shrink.
The improvements are also uneven: World Pilot trails on the Language, Robot, and Layout axes of LIBERO-Plus, and real-robot OOD success still drops by 10 to 20 points relative to ID, so the priors reduce but do not eliminate the effect of OOD shifts.
By design, the WAM and VLA are coupled only through the action loss, a modular choice that keeps either component interchangeable with stronger world models or different VLA backbones but does not pursue the tighter prior-policy co-adaptation that joint training could provide.
Each decision step also incurs an extra WAM forward pass, which limits applicability to high-frequency reactive control.
Three directions follow from these limitations: uncertainty-aware prior gating to handle WAM-coverage drops, joint WAM-VLA co-tuning to close the prior-policy loop, and prior distillation or adaptive querying to reduce the per-step overhead.


\clearpage
\acknowledgments{If a paper is accepted, the final camera-ready version will (and probably should) include acknowledgments. All acknowledgments go at the end of the paper, including thanks to reviewers who gave useful comments, to colleagues who contributed to the ideas, and to funding agencies and corporate sponsors that provided financial support.}


\bibliography{example}  

\end{document}